\documentclass{article}

\usepackage{PRIMEarxiv}
\usepackage[utf8]{inputenc} % allow utf-8 input
\usepackage[T1]{fontenc}    % use 8-bit T1 fonts
\usepackage{hyperref}       % hyperlinks
\usepackage{url}            % simple URL typesetting
\usepackage{booktabs}       % professional-quality tables
\usepackage{amsfonts}       % blackboard math symbols
\usepackage{nicefrac}       % compact symbols for 1/2, etc.
\usepackage{microtype}      % microtypography
\usepackage{lipsum}
\usepackage{fancyhdr}       % header
\usepackage{graphicx}       % graphics
\graphicspath{{media/}} 
\usepackage{amsmath,amsfonts}
\usepackage{algorithmic}
\usepackage{algorithm}
\title{Pointer Networks Trained Better via Evolutionary Algorithms}
\author{
    Muyao ZHONG\\
    Southern University of Science and Technology\\
    \And 
    Shengcai LIU\\
    Agency for Science, Technology and Research (A*STAR)
    \And 
    Bingdong LI\\
    East China Normal University
    \And 
    Haobo FU\\
    Tencent AI Lab
    \And 
    Ke TANG\\
    Southern University of Science and Technology\\
    \And
    Peng YANG\\
    Southern University of Science and Technology\\
}

\begin{document}

\maketitle

\begin{abstract}
Pointer Network (PtrNet) is a specific neural network for solving Combinatorial Optimization Problems (COPs). While PtrNets offer real-time feed-forward inference for complex COPs instances, its quality of the results tends to be less satisfactory. One possible reason is that such issue suffers from the lack of global search ability of the gradient descent, which is frequently employed in traditional PtrNet training methods including both supervised learning and reinforcement learning. To improve the performance of PtrNet, this paper delves deeply into the advantages of training PtrNet with Evolutionary Algorithms (EAs), which have been widely acknowledged for not easily getting trapped by local optima. Extensive empirical studies based on the Travelling Salesman Problem (TSP) have been conducted. Results demonstrate that PtrNet trained with EA can consistently perform much better inference results than eight state-of-the-art methods on various problem scales. Compared with gradient descent based PtrNet training methods, EA achieves up to 30.21\% improvement in quality of the solution with the same computational time. With this advantage, this paper is able to report the results of solving 1000-dimensional TSPs by training a PtrNet on the same dimensionality. This strongly suggests that the pointer network can be trained better with evolutionary algorithms for solving COPs, especially on high-dimensional instances. 
\end{abstract}

% \begin{IEEEkeywords}
% Evolutionary Algorithms, Combinatorial Optimization, Deep Neural Networks, Pointer Networks.
% \end{IEEEkeywords}

\section{Introduction}
The Pointer Network (PtrNet)\cite{ptrnet} is designed to tackle NP-hard Combinatorial Optimization Problems (COPs). PtrNet follows the typical machine learning paradigm: it is trained offline on a set of COP instances and then applied online to infer solutions for new instances from the same problem class. Due to its feed-forward inference procedure and the availability of hardware acceleration techniques\cite{liStageWiseMagnitudeBasedPruning2022}, PtrNet can solve incoming instances in seconds or less\cite{alharbiSolvingTravelingSalesman2021}. This makes PtrNet superior to  traditional mathematical programming\cite{sniedovichDynamicProgrammingFoundations2010} and meta-heuristic search methods\cite{festaBriefIntroductionExact2014} in many scenarios with real-time demands, such as Energy Management\cite{tabarEnergyManagementHybrid2019}, Unmanned Aerial Vehicle System\cite{qhouJoint2023}, and Tasks Scheduling in Cloud Computing\cite{HierRLHierarchicalReinforcement2022}. 

For this advantage, PtrNet has drawn increasing attention from the research community  recently. Its promising performance has been verified in some classical COPs like Travelling Salesman Problems (TSPs)\cite{shiPointerNetworkSolution2022}, Vehicle Routing Problems (VRPs)\cite{linDeepReinforcementLearning2022}, Job Schedule Problems (JSPs)\cite{chenLearningPerformLocal2019} and Max Clique Problems (MCPs) \cite{guPointerNetworkBased2018}. For example, when solving TSP instances with 50 nodes , the Hybrid PtrNet reduces the reference time from 1120s to 4s and obtains better results compared with OR-Tools \cite{alharbiSolvingTravelingSalesman2021}; when solving VRP instances with 30 customs and 6 charging stations, the PtrNet-based methods hierarchical GPN could achieve similar performance by consuming one-tenth less computing time of the state-of-the-art (SOTA) methods \cite{linDeepReinforcementLearning2022}.

On the other hand, one major drawback of PtrNet is the lower quality of the results compared with traditional methods, especially on larger-scale COPs, as systematically evaluated by \cite{liu2023howgood}. For example, the quality gap of result between PtrNet and the well-established solver Lin-Kernighan heuristic (LKH) \cite{helsgaunEffectiveImplementationLin2000a} degenerate significantly from 19.81\% to 41.44\% on TSP with 250 nodes and 1000 nodes \cite{maCombinatorialOptimizationGraph2019}, respectively. Although recent works have proposed more advanced models, like Policy Optimization with Multiple Optima (POMO)\cite{kwonPOMOPolicyOptimization2020},  still cannot resolve the above issue. On this basis, this paper focuses on the training method of the neural networks (NNs) for COPs, rather than the model design. We initially opted for the PtrNet over the more complex models for simplicity and generality. The straightforward architecture of PtrNet provides a clear foundation for implementing and understanding the effects of EAs in training NNs, and establishes a baseline before exploring more intricate models. 

\begin{table*}[t]
    \begin{center}
    \caption{Training and testing dimensionalities of existing PtrNet-based methods.}
    \label{table1}
    \begin{tabular}{lllcc}
        \hline
          \textbf{Problem}& \textbf{Method}  &  \textbf{Research}  &  \textbf{Training} & \textbf{Testing} \\  
          &  &  &  \textbf{Dimensionality} &  \textbf{Dimensionality} \\
        \hline
        TSP & PtrNet+SL & \cite{ptrnet} & 50 & 50 \\
        & PtrNet+RL & \cite{belloNeuralCombinatorialOptimization2016} & 100 & 100 \\
        & Graph PtrNet+RL & \cite{maCombinatorialOptimizationGraph2019} & 50 & 1000 \\
        & PtrNet+REINFORCE & \cite{koolAttentionLearnSolve2019} & 100 & 100 \\
        & PtrNet+SL & \cite{r.heHeterogeneousPointerNetwork2022} & 150 & 150 \\
        & Hybrid PtrNets & \cite{alharbiSolvingTravelingSalesman2021} & 100 & 100 \\
        \hline
        VRP & PtrNet+RL & \cite{nazariReinforcementLearningSolving2018} & 100 & 100 \\
        & PNSP-VNS & \cite{shiPointerNetworkSolution2022} & 100 & 130 \\
        & PtrNet+RL & \cite{chenLearningPerformLocal2019} & 100 & 100 \\
        & PtrNet+REINFORCE & \cite{koolAttentionLearnSolve2019} & 100 & 100 \\
        \hline
        JSP & PtrNet+RL & \cite{chenLearningPerformLocal2019} & 50 & 50 \\
        \hline
        MCP & PtrNet+SL & \cite{guPointerNetworkBased2018} & 45 & 50 \\
        \hline
    \end{tabular}
    \end{center}
\end{table*}

Existing training methods for PtrNet can be roughly classified into two categories based on how the training errors are obtained, i.e., Supervised Learning (SL) and Reinforcement Learning (RL). The major difference of these training methods lies in the origination of their error signals for optimizing the network parameters. SL adjusts the parameters by comparing the inferred solutions with the ground truth labels, i.e., the qualities of the best known solutions\cite{ptrnet}, thereby making the quality and acquisition of these labels a bottleneck for such methods. In contrast, for RL, the error signal is derived from the rewards or penalties by intermediately interacting with the environment, i.e., the evaluation function of solution qualities\cite{belloNeuralCombinatorialOptimization2016}. Despite of the error signals, both methods fundamentally relies on the gradient descent to optimize network parameters. As revealed by \cite{yangGradientGuidedEvolutionaryApproach2022}, unfortunately, a major challenge of gradient descent based methods is that their search processes are probe to be trapped by local optima. In other words, the sub-optimal results of PtrNet could possibly stem from the issue that gradient methods cannot locate a sufficiently favorable optimum within the parameter space of PtrNet. In recent years, well accepted studies have examined the potential superiority of Evolutionary Algorithms (EAs) over gradient methods for optimizing the parameters of Deep Neural Networks (DNNs) for specific problems, e.g., Deep RL for video games\cite{x.zhaoCompetitive3PlayerMahjong21} and Neural Architecture Search\cite{chengHierarchicalNeuralArchitecture2020}. A pertinent question arises that "could EA also benefit the training of PtrNets?". This paper aims to, at the first time, provide a set of sophisticated empirical evidence to answer this question.

In the framework of training DNN with EAs, EA iteratively samples a population of individuals (i.e., networks) in the parameter space based on certain randomized search operators\cite{x.zhouSurveyEvolutionaryConstruction2021}. This sampling process is essentially driven by a simulated search gradient formed by multiple evaluated individuals\cite{wierstraNaturalEvolutionStrategies2014}. As a result, EA is more likely to escape from the local optima compared to gradient descent methods, as the simulated search direction is less greedy.  Moreover, the population-based search process can further provide the possibility of exploring multiple different regions of the parameter space with the aid of population diversity\cite{NCS}. This parallel exploration ability can not only enhance the search efficiency in a way of divide-and-conquer, but also be accelerated by parallel computing\cite{yangParallelExplorationNegatively2021a}.

In order to verify the aforementioned advantages of EAs, we have designed sophisticated computational experiments on training PtrNet for TSPs. The experiments specially focus on an important training challenge. That is, existing training methods are hardly to train PtrNet effectively on 1000-dimensional instances, which are more realistic to real-world problem scales\cite{tangScalableApproachCapacitated2017}. Albeit PtrNet is able to handle varied dimensional instances on its recurrent network structure, the quality of solutions deteriorates as the difference between the testing dimensionality and training dimensionality increases, as empirically revealed by \cite{lisickiEvaluatingCurriculumLearning2020}. We also found similar phenomenon by comparing the increasing performance gap between PtrNet and LKH on solving TSPs with 250, 500, 750, 1000 dimensions, where the PtrNet is trained by gradient descent on 50 dimensional instances. %(see Fig \ref{fig:gap})
An empirical experiment shows PtrNet trained on 150 dimensions, when used to solve a problem of 1000 dimensions, has a disparity of up to 41.44\% compared to the solution obtained by LKH \cite{maCombinatorialOptimizationGraph2019}.  As suggested in these works, to solve 1000-dimensional TSPs, it is more natural and effective to train PtrNet on the same dimensionality.  

% Figure gap
% \begin{figure}[t]
% \centering
% \includegraphics[width=0.8\columnwidth]{PtrNet-EA/gap.pdf} 
% \caption{Quality Gap of the Solutions Between PtrNet Trained on TSP50 and LKH on TSP250, TSP500, TSP750 and TSP1000.}
% \label{fig:gap}
% \end{figure}

Nevertheless, Table \ref{table1} illustrates that current methods predominantly focus on various COPs  around 150 dimensions or less. While a few methods attempt solving 1000-dimensional COPs, their training instances are still no more than 150 dimensions, and the results all faced the issue of low quality. This situation remains the same for recently proposed POMO models, which underscores the importance of training and solving at similar dimensions to enhance the solution quality for high-dimensional COPs. However, this will result in more local optima in the parameter space and lead to a higher risk that the search process is trapped\cite{kawaguchiDeepLearningPoor2016}. On this basis, the challenge is a natural scenario for verifying the merits of EAs for training PtrNet.

We employed an EA called Negatively Correlated Search (NCS)\cite{NCS} to instantiate the EA training framework, denoted as PtrNet-EA. PtrNet-EA is extensively compared with eight methods, including three distinct training methods for PtrNet and five SOTA non-PtrNet methods, comprising two learning-based methods and three heuristic methods. A series of empirical studies have been conducted on TSPs and the major results are as follows:
\begin{itemize}
\item We assess the performance of learning-based algorithms on inferring 1000-dimensional TSP instances, trained with different dimensional instances within a fixed computational time budget. It shows that when the training dimensionality increase, the solution quality of the compared algorithms even degenerate by 10.57\%, while PtrNet-EA enjoys 2.68\% improvements and consistently outperforms the compared ones.
\item We test the computational time needed by all algorithms to reach the same targeted solution quality on 1000-dimensional TSP datasets. It is found that PtrNet-EA is able to consume much fewer computational time than the compared algorithms. All compared algorithms are accelerated with parallel computing units, if possible. 

\item We study the benefit of population size to the performance. By enlarging the population size up to 90, PtrNet-EA gains 4.74\% further improvement than PtrNets-EA with a population of 5.
\item We test how the diverse population of PtrNet-EA helps to solve TSPs, with making all the PtrNets in the final population as a portfolio to infer each incoming instance and recording the best result. Empirical results witness 1.99\% further improvement of PtrNet-EA, and finally achieves in total 30.21\% improvement over gradient-based PtrNet trainers, within the same acceptable computational time budget.
\end{itemize}

The contributions of this paper can be summarized as follows:
\begin{itemize}
\item This is the first work empirically assesses the advantages of training PtrNets with EA over traditional gradient based training methods by achieving in total 30.21\% improvements on 1000-dimensional COPs. This provides new insights to train neural combinatorial solvers.
\item The proposed training framework help PtrNet outperform OR-Tools and 2-opt significantly, mitigating the criticisms that neural combinatorial solvers are not competitive to existing mathematic programming based or heuristic-based solvers.
\item It is verified via extensive experiments that the advantages of training PtrNets with EA lies in the diversified search ability and the parallel computing capability, clarifying the key points of amplifying the performance of EA as the PtrNet trainer. 
\end{itemize}

The reminder of this paper is organized as follows. In section II, the related works are reviewed. The PtrNet-EA framework is presented in section III as well as its concrete version by employing the NCS algorithm. Sophisticated empirical studies are conducted in section IV to support the claims that EA is able to train PtrNet better. Conclusions are drawn in section V. 

\section{Related Work}

\noindent In this section, we review seminal works related to our study. We first discuss the application and evolution of PtrNets in COPs, highlighting their advancements and significance. Then, we delve into the application of EAs for optimizing NNs, emphasizing their potential advantages over gradient-based methods.

\subsection{PtrNets for Combinatorial Optimization Problems}
The earliest network for COPs was Hopfield network \cite{hopfieldNeuralComputationDecisions1985}. Since then, researchers have invested decades in this field to address the generalization and efficiency issues of neural combinatorial solvers. Recurrent Neural Networks (RNNs) \cite{j.wangRecurrentNeuralNetwork1996} solves the problem that Hopfield Network needs to be retrained for each instance. Unfortunately, RNNs cannot handle COPs with varied dimensions, thus their generalization ability are still restricted. With the continuous development of sequence-to-sequence (S2S) paradigm, the PtrNet, trained with SL, to learn the mapping between the input of COPs instances as a feature sequence and the output as a solution sequence. Under the S2S paradigm, the varied dimensions issue of RNNs is solved, and PtrNet ushered in a new era of neural combinatorial solvers. Thereafter, \cite{guPointerNetworkBased2018} applied PtrNet to solve Max-Cut problems(MCPs), significantly reducing the required time and computational resource for solving it with heuristic methods, demonstrating the versatility and potential of PtrNet in addressing COPs. \cite{shiPointerNetworkSolution2022} proposed a PtrNet solution pool which employs a set of PtrNets to ensure that the breadth of the search does not disappear with the iterations. \cite{r.heHeterogeneousPointerNetwork2022} introduced a heterogeneous PtrNet that adapts the encoder to the spatial-temporal feature of Traveling Officer Problems, demonstrating superior performance in capture rate and travelling distance on real-world datasets compared to existing heuristic and deep learning methods.

In addition to the aforementioned SL-based PtrNet, the PtrNet has been continuously improved under the framework of RL (PtrNet-RL) since its inception. \cite{belloNeuralCombinatorialOptimization2016} trained the PtrNet by the famous Actor-Critic algorithm \cite{i.grondmanSurveyActorCriticReinforcement2012}. They demonstrated that even trained with optimal labels, PtrNet-SL was rather poor compared to PtrNet-RL. Subsequently, \cite{nazariReinforcementLearningSolving2018} made improvements on this basis. With graph embedding to handle non-Euclidean data, \cite{maCombinatorialOptimizationGraph2019} proposed the Graph Pointer Networks (GPN) trained by RL. \cite{koolAttentionLearnSolve2019} introduced REINFORCE with greedy rollout baseline to the PtrNet-RL framework. They employed a multi-head attention mechanism to capture the relationship between the nodes in COPs, thus the results are significantly improved for small-scaled TSP. \cite{liDeepReinforcementLearning2021} leveraged PtrNet for the multi-objective TSP, and the experimental results highlight its robust generalization, fast solving speed and competitive solution quality. \cite{alharbiSolvingTravelingSalesman2021} used a hybrid PtrNet with time features to solve TSP with time windows, and their model shows superior performance over existing methods with 20, 50, and 100 nodes.

\subsection{EAs for Optimizing Neural Networks}
Beyond the listed gradient-based methods above to train PtrNets, there is a substantial body of research discussing EAs as a highly competitive alternative. These population-based algorithms offer advantages over gradient-based methods by being less susceptible to local optima, operating without the need for gradient information, and aptly handling problems without a clear objective function. Given these advantages, EAs were introduced for training NNs as early as the 1990s \cite{yaoEvolvingArtificialNeural1999}. In fact, the combination of EAs and NNs have achieved significant breakthroughs in various problems \cite{kuremotoSearchHeuristicsOptimization2020,rawalDiscoveringGatedRecurrent2020}. \cite{salimans2017evolution} proposed an Evolution Strategies (ES) to train the RL agent, in which the agent updated by the best agent that sampled from a Gaussian distribution noise of last agent. Their method is fifteen times faster than the other gradient-based methods on MuJoCo locomotion tasks. \cite{yangParallelExplorationNegatively2021a} proposed Negatively Correlated Natural ES to directly training a Deep Convolution Network with 1.7 million parameters for playing Atari games, and achieved twice acceleration over compared gradient-based methods by incorporating co-evolution mechanisms \cite{yangEvolutionaryReinforcementLearning2022}. 

Currently, there are some researchers explore to integrate EAs and neural solvers. \cite{zengEvolutionaryJobScheduling2023} uses NN to initialize the population of a genetic algorithm and further evolves the population through the genetic algorithm to obtain the final policy. \cite{zhongAcceleratingGeneticAlgorithm2022} and \cite{shaoMultiObjectiveNeuralEvolutionary2023} solve large-scaled TSP and Multi-objective TSP with EAs and PtrNets, respectively. However, due to the large parameter space of PtrNets, there has been limited research so far on directly applying EAs on PtrNets. The most related work to this paper was proposed in \cite{shaoMultiObjectiveNeuralEvolutionary2023}, where a multi-objective evolutionary algorithm based on decomposition and dominance is presented to train the PtrNet on TSP and Knapsack problems. Their works mainly focus on how to handle multiple objectives of the COPs variants or decompose large-scaled COPs instances, rather than focus on the EAs as an effective alternative of RL in training PtrNets. In this work, we at the first time thoroughly discuss the advantages mentioned in the previous section. 

\section{Methodology}
\noindent In this section, we begin by providing a concise introduction of the structure and trainable parameters of the PtrNet, and then introduce the PtrNet-EA framework. Lastly, we implement the PtrNet-EA framework by employing the NCS algorithm.

\subsection{Parameter Representation of PtrNet for EA}
\begin{figure}[!t]
\centering
\includegraphics[width=\columnwidth]{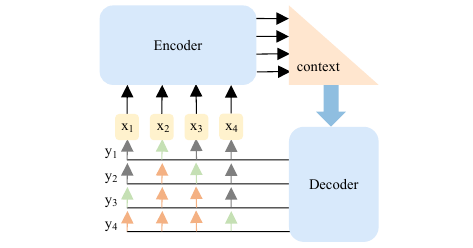} 
\caption{Flowchart of PtrNet. The green arrows denote the output pointer and the red arrows denote the pointers disabled by the masking mechanism.}
\label{ptrnet}
\end{figure}

\begin{figure}[!t]
\centering
\includegraphics[width=\columnwidth]{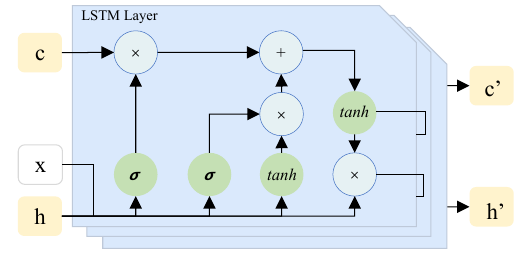} 
\caption{The structure of the encoder module in PtrNet.}
\label{encoder}
\end{figure}

\begin{figure*}[!t]
\centering
\includegraphics[width=0.8\textwidth]{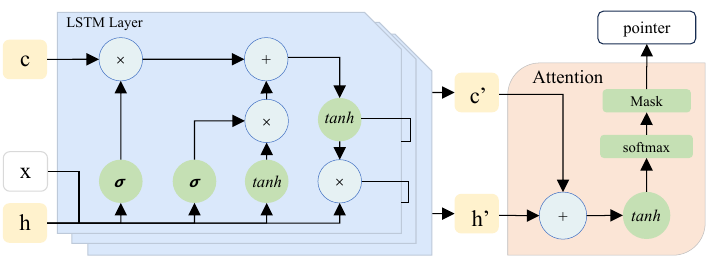} 
\caption{The structure of the decoder module in PtrNet.}
\label{decoder}
\end{figure*}

PtrNet, a specific neural architecture, comprises two Long-short Term Memory (LSTM)  modules, encoder and decoder. The encoder network sequentially processes the input sequences, which represents nodes, and coverts it into a series of latent memory states. The decoder network also maintains its latent memory state, and it employs a pointing mechanism to generate a distribution over the next chosen nodes. The parameters in both of the encoder and decoder LSTM modules are trainable, and the number of the them are listed in Table \ref{table2}, where d denotes the dimension of the hidden layer, which is 256. The workflow of PtrNet is shown in Fig.\ref{ptrnet}, and the detailed structure of the encoder and decoder of PtrNet are shown in Fig.\ref{encoder} and Fig.\ref{decoder}.

% Figure 2

\begin{table}[!t]
\centering
\caption{Trainable parameters in PtrNets}
\label{table2}

\begin{tabular}{c|c}
\hline
\textbf{Components} & \textbf{Number of parameters} \\
\hline
Encoder & $5 layers \times 4\times (d\times d)= 1310720$\\
% \hline
Decoder & $5 layers \times 4\times (d\times d)= 1310720$\\
% \hline
Attention Mechanism & $d\times d = 262144 $ \\
\hline
\end{tabular}

\end{table}

\subsection{Framework of PtrNet-EA}
An evolutionary framework for NNs is an optimization approach that draws inspiration from the natural process of evolution, aiming to optimize the structure and/or parameters of NNs through iterative generations \cite{yang2024reducing}. For neural solvers addressing COPs, their structure remains constant during training, with only the network weights requiring adjustment. Based on this, in this work the evolutionary framework is employed to optimize the parameters of the PtrNet while keeping the architecture fixed. The following workflow describes how EAs optimize the parameters, each individual in the population denotes the parameters of a PtrNet.
\begin{enumerate}
    \item \textbf{Initialization.} Generate an initial population of candidate PtrNets, each with a distinct  parameters. The population size is predefined, and the architectures is based on previous works.
    \item \textbf{Fitness Evaluation.} Evaluate the performance of the individuals using a predefined objective function.
    \item \textbf{Reproduction.} The search operators of EAs, e.g., selector, crossover, and mutation, are used for generating the next population.
    \item \textbf{Termination Condition.} Repeat step 2 and 3 until a stopping criterion is met, e.g., a predefined time budget, reaching a performance threshold or detecting convergence.
\end{enumerate}

Upon completion of the evolutionary process, the PtrNet with the highest fitness is considered the optimized policy for inferring COPs. 

\subsection{Concrete EA-based Training Algorithm}
In this paper, we choose NCS to instantiate the PtrNet-EA. NCS is a type of EA with explicitly enhanced diversity and has been applied to solve numerous practical problems \cite{yangParallelExplorationNegatively2021a,yangEvolutionaryReinforcementLearning2022}. NCS consists of multiple parallel search processes. Each search process is driven by a Gaussian mutation operator \cite{xinyaoEvolutionaryProgrammingMade1999}, which generates the new individual $x_i^\prime$ by sampling from $\mathcal{N}\left(x_i,\sigma_i\right)$, where $\sigma_i$ is the standard deviation of the Gaussian distribution $\mathcal{N}$ and represents the search step-size. The individuals are expected to be with high quality and distant from each other to form a parallel exploration of multiple regions of the parameter space. In NCS, the distance is defined as Bhattacharyya distance \cite{t.kailathDivergenceBhattacharyyaDistance1967} by Eq.(1), where $p_i$ represents the distributions corresponding to other individuals. NCS forces the population seeking different optimal areas by trading off between the distance and the objective fitness values with an adaptive parameter $\lambda_t$. The $\sigma_i$ is updated based on the 1/5 success rule \cite{beyerEvolutionStrategiesComprehensive2002b}. That is, $\sigma_i$ will be increased to enhance the exploration if 20\% of the offspring $x_i^\prime$ is better than its parent $x_i$ within $t_e$ iteration; otherwise, it will be reduced for favoring exploitation, according to Eq.(2), where $c_i$ counts the replaced individuals. The pseudo-code is given in Algorithm 
\ref{alg1}
, and at the end of the algorithm, the final population obtains all the PtrNet individuals with good performance and diversity. For more details of NCS, please refer to \cite{NCS}.

\small
\begin{equation}
Corr\left(p_i\right)={\mathbf{min}}_j{\{D_B(p_i,p_j)|j\neq i\}}
\end{equation}

\begin{equation}
\sigma_i = \begin{cases} 
\sigma_i / 0.99, \; \text{if } c_i/t_e \leq 0.2 ; \\
0.99\sigma_i, \; \text{otherwise}.
\end{cases}
\end{equation}

% \normalsize
\begin{algorithm}[tb]
\caption{PtrNet-EA}
\label{alg1}
\textbf{Input}: $T_{max}$,$t_e$,$N$\\
\textbf{Output}: Last population
\begin{algorithmic}[1] %[1] enables line numbers
\STATE Randomly generate an initial population of N PtrNets.
\STATE Evaluate the PtrNets by objective function $f$.
\STATE Set $t\leftarrow 0$.
\WHILE{$t< T_{max}$}
\STATE Set $\lambda_t \leftarrow \mathcal{N}(1, 0.1-0.1t/T_{max})$ 
\FOR{ $i=1$ to $N$}
\STATE Generate a new PtrNet $x_i^\prime\leftarrow\mathcal{N}(x_i,\sigma_i)$
\STATE Compute $f(x^\prime_i)$ and $Corr(p_i^\prime)$
\STATE $x_i\leftarrow x_i^\prime$ if $f(x^\prime_i)$ / $Corr(p_i^\prime)<\lambda_t$
\ENDFOR
\STATE $t\leftarrow t+1$
\IF{$mod(t,t_e)=0$}
\STATE Update every $\sigma_i$ according to Eq.(2)
\ENDIF
\ENDWHILE
\STATE \textbf{return} solution
\end{algorithmic}
\end{algorithm}

\section{Empirical Studies}
\noindent This section assesses the potential of PtrNet-EA in the following aspects:
\begin{itemize}
    \item Whether EAs can train the PtrNets better than traditional gradient-based methods within an acceptable training time budget?
    \item How would the parallel computing accelerate EAs in training pointer network?
    \item How does the population diversity facilitate the performance of pointer network?
\end{itemize}

\subsection{Travelling Salesman Problem}
In this paper, the TSP is selected as the benchmark problems as it is commonly tested as a typical NP-hard combinatorial optimization problems \cite{ptrnet}. In general, given a set of nodes in a graph and their coordinates, the goal of TSP is to find the shortest route that visits each node exactly once and returns to the starting node. 

Formally, let  $X=\{X_1,X_2,\dots , X_n\}$ be $n$ nodes in the TSP instance. The goal is to optimize a permutation  over the nodes that minimises the path length
\begin{equation}
     L(\pi, X) = \sum\nolimits_{i=1}^{n} \left\| X_{\pi(i)} - X_{\pi(i+1)} \right\|_2\\
\end{equation}
   
subject to:
\small
$$
    \pi(1)=\pi(n+1),
$$
$$
    \pi(i)\neq \pi(j),
$$
$$
    \pi(i)\in \{ 1,2,\dots,n\}
$$
\normalsize
Where $\sigma\left(i\right)$ is the index of node $i$ in the path sequence. Each node $X_i$ is represented by the coordinates where it is located. That is $X_i=(x_i,x_2,\ldots,x_d)$, where $d$ is the dimension of the coordinates. In this paper, two-dimensional Euclidean coordinates are used for the representation, or $d=2$. 

For the TSP instances, existing real-world benchmarks have been found insufficient for training PtrNets in terms of number of instances \cite{maCombinatorialOptimizationGraph2019}. Therefore, following the method by \cite{maCombinatorialOptimizationGraph2019}, we synthesize three sets of TSP instances with 100 nodes, 500 nodes, and 1000 nodes, denoted as TSP100, TSP500, and TSP1000, respectively. Each set consists of 1 million training instances and 2000 testing instances, which are sampled from the same instance distribution by randomly distributing the x, y coordinates of the nodes in the real-valued interval of $\left[0,1\right]^2$.

\subsection{Algorithms Settings}
Suggested by \cite{mazyavkinaReinforcementLearningCombinatorial2021}, three types of well-established combinatorial solvers are selected as the compared algorithms. The first type contains the PtrNet-based methods, including the original PtrNet trained by SL (PtrNet-SL), the Graph Pointer Network with 2-opt (GPN-RL) \cite{maCombinatorialOptimizationGraph2019}, the pointer network trained by RL (PtrNet-RL) \cite{belloNeuralCombinatorialOptimization2016}. The second type includes three non-PtrNet learning-based solvers, i.e., POMO\cite{kwonPOMOPolicyOptimization2020}, the Structure to Vector Deep Q-learning (S2V-DQN) \cite{khalilLearningCombinatorialOptimization2017} and Variable Strategy Reinforced LKH (VSR-LKH) \cite{zhengCombiningReinforcementLearning2021}. S2V-DQN is a deep learning architecture over graphs, and VSR-LKH is a hybrid of RL and LKH. The third type consists of the heuristic search based SOTA methods, i.e., the 2-opt \cite{crama2005local}, the Google OR-Tools with saving algorithm, and Focused Ant Colony Optimization (FACO)\cite{skinderowiczImprovingAntColony2022}.

By comparing with the first two types, it aims to show the learning ability of the proposed PtrNet-EA. By comparing with the third type of methods, it emphasizes the superiority of PtrNet-EA over the traditional solvers within an acceptable time budget. The parameters and implementations of the above compared algorithms exactly follow the source codes released by \cite{maCombinatorialOptimizationGraph2019}. The hyperparameters of PtrNet-EA are listed in Table \ref{table3}. The embedding size, weight size, and batch size are set to the same with \cite{ptrnet}. There are approximately 2.9M weights to optimize. The population size is set to 5 while comparing with SOTAs, and later set to 50 and 90 for parallel assessment. 

\begin{table}[t]
\centering
\caption{Hyperparameters used in PtrNet-EA.}
\label{table3}
\begin{tabular}{c|ll}
\hline
\textbf{Component} & \textbf{Hyperparameter} & \textbf{Value} \\
\hline
PtrNet & Embedding Size & 32 \\
                        & Hidden State Size & 256 \\
                        & Layer Number & 5 \\
                        & Batch Size & 256 \\
\hline
NCS & Max Time Steps & 8000 \\
                     & Update frequency & 10 \\
                     & Population Size & [5, 50, 90] \\
\hline
\end{tabular}

\end{table}

\subsection{Experimental Settings}
This work aims to study how to train PtrNets with 1000-dimensionality TSP efficiently. We set the time budget to 500 minutes for all the algorithms for fairness. Indeed, any other time budgets larger than 500 minutes do not mean unacceptable. We simply approximate it around one working day of 8 hours for the consideration of real-world activities. More specifically, for learning-based solvers, the training phase is set to 500 minutes, as the testing phase is usually very efficient and thus the testing time can be omitted. For the heuristic search-based methods, we allow them to solve the instances in 500 minutes. Unless specified, all experimental results are the average of three independent runs on the target TSP testing set each with 2000 problem instances. All experiments are conducted on a workstation with 320GB RAM and 96 cores (4.0GHz, 71.5MB L3 Cache), running Ubuntu 20.04\footnote{The code can be found https://github.com/mythezone/PtrNetEA.git}. The experiments are divided into four Groups and will be described in detail.

\textbf{Group I.} We compare the proposed PtrNet-EA with the 9 chosen algorithms. All the learning-based solvers are first trained with 1 million training instances of TSP100, and then tested on 2000 testing instances of TSP100, TSP500, and TSP1000, respectively.  The 2-opt, OR-Tools and FACO are directly used to solve the 2000 testing instances, as they do not involve the training step. The output path lengths of 2000 testing instances are averaged as the final performance of each algorithm. 

\textbf{Group II.} A question naturally arises why existing works do not train the networks at the same scale with the testing instances. To clarify this, we train all the learning-based solvers on the training instances of TSP1000 and test them on the testing instances of TSP1000. In order to further analyze the superiority of PtrNet-EA on TSP1000, we use the testing performance of PtrNet-EA trained in the above-mentioned 500 minutes time budget as the baseline and run the compared algorithms with a 10x time budget. For every 50 minutes, we record the performance of compared algorithms on the testing instances. If the performance of a compared algorithm reaches the baseline, we record the consumed running time for that algorithm. Otherwise, we record the final performance of that algorithm stopped at 5000 minutes. This group of comparisons verifies whether EAs could help pointer networks achieve better performance in terms of running time on larger problems.

\textbf{Group III.} To verify how the parallel computing accelerates EAs in training PtrNets, the population size N of PtrNet-EA is set to 5, 50 and 90, respectively. For each of them, N processors are allocated to parallelize the training of individual policies, and the training time budget is also set to 500 minutes. By this means, it is expected to see that: 1) all the three algorithms will run for the same iterations; 2) due to the larger population sampled in each iteration, the PtrNet-EA with larger population size will achieve better results. If this empirically holds, it is promising to say that by utilizing more parallel computing resources, EAs can train the PtrNets even better within the same time budget.

\textbf{Group IV.} By treating the final population of evolved PtrNets as a policy portfolio, the complementary strengths of the individual policies can be leveraged to further improve the performance on COP instances. We verify this by comparing the best policy with policy portfolio, denoted as PtrNet-EA-P, on TSP1000.

\subsection{Results and Analysis}
\begin{figure*}[!t]
\centering
\includegraphics[width=\textwidth]{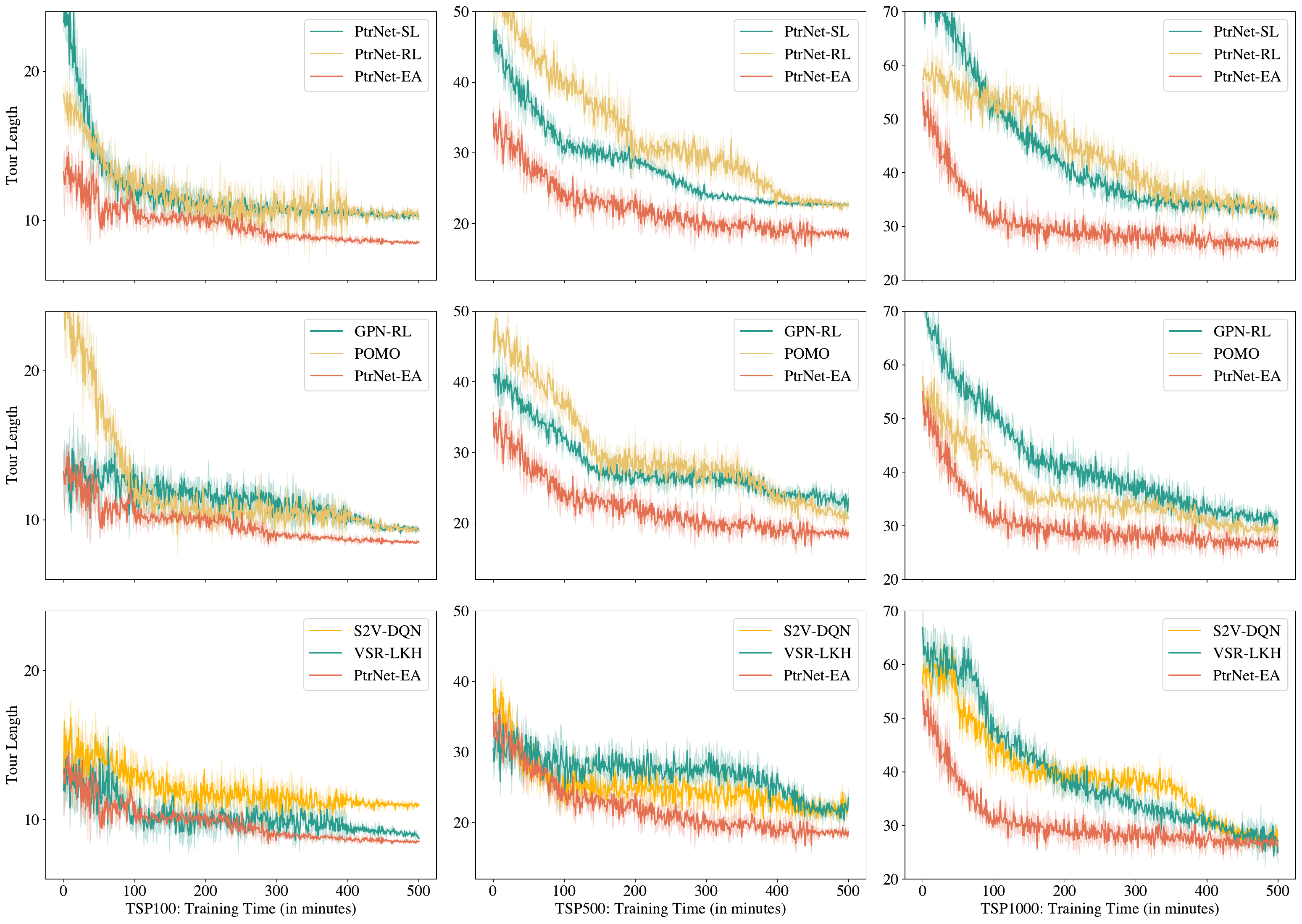} 
\caption{A comparison of the convergence curves during the training process of other methods against PtrNet-EA. The  three columns correspond to the training conducted on TSP100, TSP500, and TSP1000, respectively. And the three rows depict three groups of different compared algorithms. The vertical axis represents the Tour Length for the current batch on the Training set.
}
\label{fig-train}
\end{figure*}

\textbf{Group I.} 
As illustrated in Fig.\ref{fig-train}, throughout the training process, the PtrNet-EA not only demonstrates a faster convergence rate but also achieves the best performance on the training set among all the evaluated methods. 

Comparing the training curve across three columns in Fig.\ref{fig-train}, it is observed that in TSP100 and TSP500, PtrNet-EA does not show a significant difference in convergence trends compared to other methods. However, as the problem size increases to TSP1000, the disparity between PtrNet-EA and other approaches becomes more pronounced. 
The first row of comparative reveals that, across TSP100, TSP500, TSP1000, the EA trains PtrNet more effectively, when compared with SL and RL. Notably, since the PtrNet-SL method employs approximate solutions as labels for training, it converges faster than the RL method in some cases. Comparing the second and third rows, on the training set, the EA achieves better training performance.
This highlights the effectiveness of PtrNet-EA in terms of both learning efficiency and solution quality within the given training scope.

When compared with PtrNet-SL, PtrNet-EA has 18.35\%, 18.70\% and 21.61\% improvements on TSP100, TSP500 and TSP1000, respectively. When compared with PtrNet-RL, PtrNet-EA has at least 11.95\%, 17.23\% and 10.91\% improvements on TSP100, TSP500 and TSP1000,  respectively. These comparisons show that the PtrNet-EA has a great learning ability. 
When compared with other SOTA methods, the results have been improved by at least 4.40\%, 4.27\% and 2.22\% (FACO) on TSP100, TSP500, TSP1000, respectively. The result means that PtrNet-EA is able to achieve better solution in time-constrained tasks. The detailed results are shown in Table \ref{table4}.

\begin{table}[t]
\centering
\caption{Comparison of methods trained on TSP100 and Tested on TSP100, TSP500 and TSP1000, by $Avg\pm Std$.}
\label{table4}
\begin{tabular}{r|rrr}
\hline
\textbf{Methods} & \textbf{TSP100} & \textbf{TSP500} & \textbf{TSP1000} \\
\hline
PtrNet-SL & $10.29 \pm 0.60$ & $22.61 \pm 0.85$ & $33.89 \pm 0.36$ \\
% \hline
PtrNet-RL & $10.60 \pm 0.98$ & $22.21 \pm 0.92$ & $31.44 \pm 0.53$ \\
% \hline
GPN-RL & $9.54 \pm 0.45$ & $22.40 \pm 0.24$ & $29.82 \pm 0.27$ \\
% \hline
POMO & $9.44 \pm 0.89$ & $20.78 \pm 0.42$ & $29.94 \pm 0.90$ \\

S2V-DQN & $11.09 \pm 1.12$ & $22.21 \pm 0.30$ & $27.66 \pm 0.36$ \\
% \hline
VSR-LKH & $8.94 \pm 0.19$ & $21.22 \pm 0.56$ & $27.30 \pm 0.43$ \\
% \hline
OR-Tools & $9.41 \pm 0.09$ & $21.26 \pm 0.82$ & $29.50 \pm 0.85$ \\
% \hline
2-opt & $9.02 \pm 0.67$ & $20.16 \pm 0.26$ & $29.21 \pm 0.41$ \\
% \hline
FACO & $8.79 \pm 0.16$ & $19.20 \pm 0.21$ & $27.17 \pm 0.52$ \\
% \hline
\textbf{PtrNet-EA} &  \textbf{8.40 $\pm$ 0.13} & \textbf{18.38 $\pm$ 0.57} & \textbf{26.57 $\pm$ 0.56} \\
\hline
\end{tabular}

\end{table}

Next, we compared all the learning-based methods trained on TSP1000 and tested on TSP1000. Not surprisingly, PtrNet-EA still maintains the leading position in this comparison. As Table \ref{table5} shows, some of the other methods have performance loss in this experiment especially the RL-based ones. However, PtrNet-EA benefits from the high-dimensional datasets, and the quality of the solution is improved by 1.74\% in comparison with training on TSP100.

Integrating the results of these two comparative experiments in Group I, it can be inferred that with the premise of a reasonable time budget, EAs demonstrate superior efficacy in searching the parameter space to identify optimal solutions, compared to other methods. Furthermore, when compared to traditional methods, PtrNet-EA achieves better results within the same computational time budget. Therefore, we could draw the conclusion that under the constraint of limited computational resources, the EA method  trains neural networks more effectively and yields superior solutions.

\begin{table}[t] 
\small
\centering
\caption{The solution quality of learning-based models trained on TSP100 and TSP1000, 
and Tested on TSP1000, by $Avg\pm Std$.
}
\label{table5}
\begin{tabular}{r|rr}
\hline 
\textbf{Methods} & \textbf{Trained on TSP100} & \textbf{Trained on TSP1000} \\
\hline
PtrNet-SL & 33.89 $\pm$ 0.36 & 32.75 $\pm$ 0.72 \\
% \hline
PtrNet-RL & 31.44 $\pm$ 0.53 & 31.98 $\pm$ 0.80 \\
% \hline
GPN-RL & 29.82 $\pm$ 0.27 & 29.29 $\pm$ 0.81 \\
% \hline
POMO & 28.61 $\pm$ 0.71 & 32.69 $\pm$ 0.23 \\

S2V-DQN & 27.66 $\pm$ 0.36 & 30.62 $\pm$ 0.70 \\
% \hline
VSR-LKH & 27.30 $\pm$ 0.43 & 32.35 $\pm$ 0.20 \\
% \hline
\textbf{PtrNet-EA} & \textbf{26.57 $\pm$ 0.56} & \textbf{25.86 $\pm$ 0.58} \\
\hline
\end{tabular}

\end{table}

\begin{figure}[!t]
\centering
\includegraphics[width=\columnwidth]{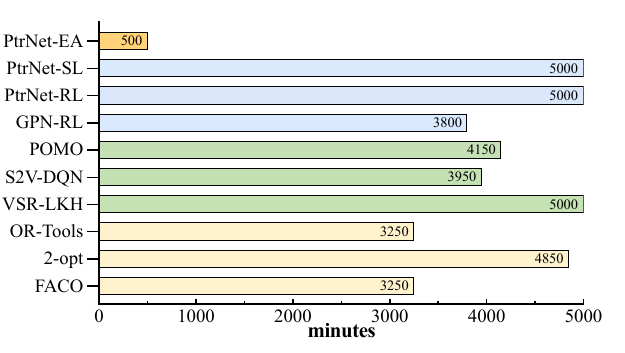} 
\caption{10x time budget for compared methods training on TSP1000 to reach the PtrNet-EA’s performance training in 500 minutes.
}
\label{fig4}
\end{figure}

\textbf{Group II.} The results are shown in Fig.\ref{fig4}. It is obvious that the compared methods need over 6.5x the time budget to achieve the same performance of PtrNet-EA. PtrNet-SL and PtrNet-RL even cannot achieve the same performance of PtrNet-EA in 10x the time budget.

We posit that the observed results are attributable to two primary factors. Firstly, the population-based search characteristic of EAs aligns well with the highly parallelized architecture of modern computers, enabling more efficient utilization of hardware for acceleration compared to other algorithms. Secondly, the selection of an EA that is well-matched to the problem (in our case, training neural networks) results in a more effective task completion.

NCS explicitly incorporates the distance between individuals in the population into the fitness function and adaptively adjusts the step size during the searching process. This approach emphasizes population diversity and the characteristic of obtaining offspring through sampling distributions, which makes NCS particularly suitable for searching high-dimensional neural network parameters. It also suggests that EAs with good diversity maintenance techniques should be able to train PtrNet well.

\begin{figure}[!t]
\centering
\includegraphics[width=\columnwidth]{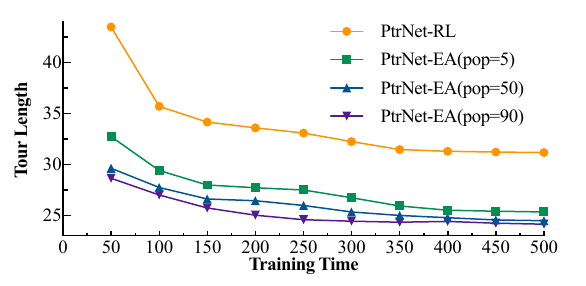} 
\caption{Convergence curve of PtrNet-EA with different population size and PtrNet-RL.}
\label{fig5}
\end{figure}

\textbf{Group III.} Based on the parallel nature of EAs, we believe that a larger population can achieve better search results. Fig.\ref{fig5} shows the effect of population size to PtrNet-EA. From the convergence curve, we can clearly figure out that larger population size could reach the specified performance faster, and within a given time, larger one performs better, which confirmed our conjecture. By removing the diversity control element from PtrNet-NCS(pop5), the PtrNet-EA(noDiv5)  method got a huge drop of 26.8\% (see Table \ref{table6}), which also verified the importance of population diversity in PtrNet-EA indirectly.  It is worth mentioning that in parallel condition, there is almost no extra training time in PtrNet-EA. 

This set of experiments comprehensively demonstrates the superiority of EAs in training neural networks. Under the condition of maintaining diversity among individuals in the population, a larger population size signifies a more effective exploration of the parameter space. This not only implies the potential for a time-efficient search by trading resources for time, achieving the same solution quality in less time. It also suggests the possibility of obtaining better results within the same time budget by further increasing computational resources. In the context of distributed  computation architectures, where abundant and cost-effective computing resources are readily available, a method that can fully leverage these resources will offer more superior solutions for users with specific needs.

\begin{figure}[!t]
\centering
\includegraphics[width=0.8\columnwidth]{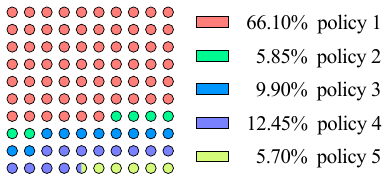} 
\caption{Diversity of the policy portfolio (pop5).}
\label{diversity}
\end{figure}

\begin{table}[!t]
\centering
\caption{Performance difference between PtrNet-EA and PtrNet-EA-P.
}
\label{table6}
\begin{tabular}{l|cc}
\hline
\textbf{Method} & \textbf{PtrNet-EA} & \textbf{PtrNet-EA-P} \\
\hline
PtrNet-EA(noDiv5) & 32.12 & 31.87 \\
% \hline
PtrNet-EA(pop5) & 25.33 & 25.12 \\
% \hline
PtrNet-EA(pop50) & 24.49 & 24.21 \\
% \hline
\textbf{PtrNet-EA(pop90)} & \textbf{24.13} & \textbf{23.65} \\
\hline
\end{tabular}

\end{table}

\textbf{Group IV.} Last but not least, the policy portfolio can also make contributions to the testing performance. When we use all policies in the last population as a portfolio, each policy has the potential to obtain the optimal solution on specific problem instances. As shown in Fig.\ref{diversity}, 34\% best solutions of the 2000 testing instances are obtained by the policies in the portfolio other than the best policy. As much as 1.99\% improvement is introduced by the policy portfolio (see Table \ref{table6}), and the overall improvement reached an astonishing 30.21\% compared with the original PtrNet-SL. 

While other stochastic methods may also employ portfolio polices to solve problems multiple times, retain the best solution to enhance performance without adding extra waiting time due to parallel execution. The solutions obtained might lack diversity, which means the results of each run could be very similar.  In contrast, the EAs inherently comprises  population with multiple individuals. It not only avoids additional time budget but also utilizes the diversity within these individuals with minimal computational resources required for infering the solutions. This approach enables the generation of substantially varied computational outcomes, thereby potentially improving the quality of the solutions.

\section{Conclusion}
\noindent In this paper, we delved deeply into the performance constrains observed in the PtrNets when solving COPs. Recognizing the limitations of gradient descent methods, primarily their susceptibility to being trapped in local optima, our study postulated the potential effectiveness of EAs as a competitive alternative. To verify this, we systematically validated these through sophisticated four groups of empirical experiments, and the results consistently demonstrated the superior capability of EAs in training PtrNets.

Therefore, our work provides a new insight to both fields of EAs and neural combinatorial solvers, which indicating that EAs could train PtrNets more proficiently than gradient-based methods. This suggests that for applications that leverage PtrNet in solving large-scale COPs, utilizing EAs for training should be seriously considered.

Moreover, while our current research focused on the PtrNet, it is reasonable to induce that the other neural combinatorial solvers can also benefit from similar evolutionary training framework. In the future, how to train more advanced solvers with EAs will be studied as their network models appear to be more complex. 

\bibliographystyle{unsrt}  
\bibliography{reference}

\begin{thebibliography}{10}

\bibitem{ptrnet}
Oriol Vinyals, Meire Fortunato, and Navdeep Jaitly.
\newblock Pointer {{Networks}}.
\newblock In {\em Advances in {{Neural Information Processing Systems}}}, volume~28. {Curran Associates, Inc.}, 2015.

\bibitem{liStageWiseMagnitudeBasedPruning2022}
Guiying Li, Peng Yang, Chao Qian, Richang Hong, and Ke~Tang.
\newblock Stage-{{Wise Magnitude-Based Pruning}} for {{Recurrent Neural Networks}}.
\newblock {\em IEEE Transactions on Neural Networks and Learning Systems}, pages 1--15, 2022.

\bibitem{alharbiSolvingTravelingSalesman2021}
Majed~G. Alharbi, Ahmed Stohy, Mohammed Elhenawy, Mahmoud Masoud, and Hamiden~Abd {El-Wahed Khalifa}.
\newblock Solving {{Traveling Salesman Problem}} with {{Time Windows Using Hybrid Pointer Networks}} with {{Time Features}}.
\newblock {\em Sustainability}, 13(22):12906, January 2021.

\bibitem{sniedovichDynamicProgrammingFoundations2010}
Moshe Sniedovich.
\newblock {\em Dynamic Programming: Foundations and Principles}.
\newblock {CRC press}, 2010.

\bibitem{festaBriefIntroductionExact2014}
P.~Festa.
\newblock A brief introduction to exact, approximation, and heuristic algorithms for solving hard combinatorial optimization problems.
\newblock In {\em 2014 16th {{International Conference}} on {{Transparent Optical Networks}} ({{ICTON}})}, pages 1--20, July 2014.

\bibitem{tabarEnergyManagementHybrid2019}
Vahid~Sohrabi Tabar, Saeid Ghassemzadeh, and Sajjad Tohidi.
\newblock Energy management in hybrid microgrid with considering multiple power market and real time demand response.
\newblock {\em Energy}, 174:10--23, May 2019.

\bibitem{qhouJoint2023}
Qiushuo Hou, Yunlong Cai, Qiyu Hu, Mengyuan Lee, and Guanding Yu.
\newblock Joint resource allocation and trajectory design for multi-uav systems with moving users: Pointer network and unfolding.
\newblock {\em IEEE Transactions on Wireless Communications}, 22(5):3310--3323, 2023.

\bibitem{HierRLHierarchicalReinforcement2022}
{Y. Guan}, {Y. Liu}, {Y. Li}, and {X. Xu}.
\newblock {{HierRL}}: {{Hierarchical Reinforcement Learning}} for {{Task Scheduling}} in {{Distributed Systems}}.
\newblock In {\em 2022 {{International Joint Conference}} on {{Neural Networks}} ({{IJCNN}})}, pages 1--8, July 2022.

\bibitem{shiPointerNetworkSolution2022}
Chenxu Shi.
\newblock Pointer {{Network Solution Pool}} : {{Combining Pointer Networks}} and {{Heuristics}} to {{Solve TSP Problems}}.
\newblock In {\em 2022 3rd {{International Conference}} on {{Computer Vision}}, {{Image}} and {{Deep Learning}} \& {{International Conference}} on {{Computer Engineering}} and {{Applications}} ({{CVIDL}} \& {{ICCEA}})}, pages 1236--1242, May 2022.

\bibitem{linDeepReinforcementLearning2022}
Bo~Lin, Bissan Ghaddar, and Jatin Nathwani.
\newblock Deep {{Reinforcement Learning}} for the {{Electric Vehicle Routing Problem With Time Windows}}.
\newblock {\em IEEE Transactions on Intelligent Transportation Systems}, 23(8):11528--11538, August 2022.

\bibitem{chenLearningPerformLocal2019}
Xinyun Chen and Yuandong Tian.
\newblock Learning to {{Perform Local Rewriting}} for {{Combinatorial Optimization}}.
\newblock In {\em Advances in {{Neural Information Processing Systems}}}, volume~32. {Curran Associates, Inc.}, 2019.

\bibitem{guPointerNetworkBased2018}
Shenshen Gu and Yue Yang.
\newblock A {{Pointer Network Based Deep Learning Algorithm}} for the {{Max-Cut Problem}}.
\newblock In Long Cheng, Andrew Chi~Sing Leung, and Seiichi Ozawa, editors, {\em Neural {{Information Processing}}}, Lecture {{Notes}} in {{Computer Science}}, pages 238--248, {Cham}, 2018. {Springer International Publishing}.

\bibitem{liu2023howgood}
Shengcai Liu, Yu~Zhang, Ke~Tang, and Xin Yao.
\newblock How good is neural combinatorial optimization? a systematic evaluation on the traveling salesman problem.
\newblock {\em IEEE Computational Intelligence Magazine}, 18(3):14--28, 2023.

\bibitem{helsgaunEffectiveImplementationLin2000a}
Keld Helsgaun.
\newblock An effective implementation of the {{Lin}}\textendash{{Kernighan}} traveling salesman heuristic.
\newblock {\em European Journal of Operational Research}, 126(1):106--130, October 2000.

\bibitem{maCombinatorialOptimizationGraph2019}
Qiang Ma, Suwen Ge, Danyang He, Darshan Thaker, and Iddo Drori.
\newblock Combinatorial optimization by graph pointer networks and hierarchical reinforcement learning.
\newblock {\em arXiv preprint arXiv:1911.04936}, 2019.

\bibitem{kwonPOMOPolicyOptimization2020}
Yeong-Dae Kwon, Jinho Choo, Byoungjip Kim, Iljoo Yoon, Youngjune Gwon, and Seungjai Min.
\newblock {{POMO}}: {{Policy Optimization}} with {{Multiple Optima}} for {{Reinforcement Learning}}.
\newblock In {\em Advances in {{Neural Information Processing Systems}}}, volume~33, pages 21188--21198. {Curran Associates, Inc.}, 2020.

\bibitem{belloNeuralCombinatorialOptimization2016}
Irwan Bello, Hieu Pham, Quoc~V. Le, Mohammad Norouzi, and Samy Bengio.
\newblock Neural combinatorial optimization with reinforcement learning.
\newblock {\em arXiv preprint arXiv:1611.09940}, 2016.

\bibitem{koolAttentionLearnSolve2019}
Wouter Kool, Herke {van Hoof}, and Max Welling.
\newblock Attention, {{Learn}} to {{Solve Routing Problems}}!, February 2019.

\bibitem{r.heHeterogeneousPointerNetwork2022}
{R. He}, {X. Xiao}, {Y. Kang}, {H. Zhao}, and {W. Shao}.
\newblock Heterogeneous {{Pointer Network}} for {{Travelling Officer Problem}}.
\newblock In {\em 2022 {{International Joint Conference}} on {{Neural Networks}} ({{IJCNN}})}, pages 1--8, July 2022.

\bibitem{nazariReinforcementLearningSolving2018}
MohammadReza Nazari, Afshin Oroojlooy, Lawrence Snyder, and Martin Takac.
\newblock Reinforcement {{Learning}} for {{Solving}} the {{Vehicle Routing Problem}}.
\newblock In {\em Advances in {{Neural Information Processing Systems}}}, volume~31. {Curran Associates, Inc.}, 2018.

\bibitem{yangGradientGuidedEvolutionaryApproach2022}
{S. Yang}, {Y. Tian}, {C. He}, {X. Zhang}, {K. C. Tan}, and {Y. Jin}.
\newblock A {{Gradient-Guided Evolutionary Approach}} to {{Training Deep Neural Networks}}.
\newblock {\em IEEE Transactions on Neural Networks and Learning Systems}, 33(9):4861--4875, September 2022.

\bibitem{x.zhaoCompetitive3PlayerMahjong21}
{X. Zhao} and {S. B. Holden}.
\newblock Towards a {{Competitive}} 3-{{Player Mahjong AI}} using {{Deep Reinforcement Learning}}.
\newblock In {\em 2022 {{IEEE Conference}} on {{Games}} ({{CoG}})}, pages 524--527, 21.

\bibitem{chengHierarchicalNeuralArchitecture2020}
Xuelian Cheng, Yiran Zhong, Mehrtash Harandi, Yuchao Dai, Xiaojun Chang, Hongdong Li, Tom Drummond, and Zongyuan Ge.
\newblock Hierarchical {{Neural Architecture Search}} for {{Deep Stereo Matching}}.
\newblock In H.~Larochelle, M.~Ranzato, R.~Hadsell, M.~F. Balcan, and H.~Lin, editors, {\em Advances in {{Neural Information Processing Systems}}}, volume~33, pages 22158--22169. {Curran Associates, Inc.}, 2020.

\bibitem{x.zhouSurveyEvolutionaryConstruction2021}
{X. Zhou}, {A. K. Qin}, {M. Gong}, and {K. C. Tan}.
\newblock A {{Survey}} on {{Evolutionary Construction}} of {{Deep Neural Networks}}.
\newblock {\em IEEE Transactions on Evolutionary Computation}, 25(5):894--912, October 2021.

\bibitem{wierstraNaturalEvolutionStrategies2014}
Daan Wierstra, Tom Schaul, Tobias Glasmachers, Yi~Sun, Jan Peters, and J{\"u}rgen Schmidhuber.
\newblock Natural evolution strategies.
\newblock {\em The Journal of Machine Learning Research}, 15(1):949--980, 2014.

\bibitem{NCS}
Ke~Tang, Peng Yang, and Xin Yao.
\newblock Negatively correlated search.
\newblock {\em IEEE Journal on Selected Areas in Communications}, 34(3):542--550, 2016.

\bibitem{yangParallelExplorationNegatively2021a}
Peng Yang, Qi~Yang, Ke~Tang, and Xin Yao.
\newblock Parallel exploration via negatively correlated search.
\newblock {\em Frontiers of Computer Science}, 15(5):155333, July 2021.

\bibitem{tangScalableApproachCapacitated2017}
Ke~Tang, Juan Wang, Xiaodong Li, and Xin Yao.
\newblock A {{Scalable Approach}} to {{Capacitated Arc Routing Problems Based}} on {{Hierarchical Decomposition}}.
\newblock {\em IEEE Transactions on Cybernetics}, 47(11):3928--3940, November 2017.

\bibitem{lisickiEvaluatingCurriculumLearning2020}
Michal Lisicki, Arash Afkanpour, and Graham~W. Taylor.
\newblock Evaluating {{Curriculum Learning Strategies}} in {{Neural Combinatorial Optimization}}, November 2020.

\bibitem{kawaguchiDeepLearningPoor2016}
Kenji Kawaguchi.
\newblock Deep {{Learning}} without {{Poor Local Minima}}.
\newblock In {\em Advances in {{Neural Information Processing Systems}}}, volume~29. {Curran Associates, Inc.}, 2016.

\bibitem{hopfieldNeuralComputationDecisions1985}
J.~J. Hopfield and D.~W. Tank.
\newblock ``{{Neural}}'' computation of decisions in optimization problems.
\newblock {\em Biological Cybernetics}, 52(3):141--152, July 1985.

\bibitem{j.wangRecurrentNeuralNetwork1996}
{J. Wang}.
\newblock A recurrent neural network for solving the shortest path problem.
\newblock {\em IEEE Transactions on Circuits and Systems I: Fundamental Theory and Applications}, 43(6):482--486, June 1996.

\bibitem{i.grondmanSurveyActorCriticReinforcement2012}
{I. Grondman}, {L. Busoniu}, {G. A. D. Lopes}, and {R. Babuska}.
\newblock A {{Survey}} of {{Actor-Critic Reinforcement Learning}}: {{Standard}} and {{Natural Policy Gradients}}.
\newblock {\em IEEE Transactions on Systems, Man, and Cybernetics, Part C (Applications and Reviews)}, 42(6):1291--1307, November 2012.

\bibitem{liDeepReinforcementLearning2021}
Kaiwen Li, Tao Zhang, and Rui Wang.
\newblock Deep {{Reinforcement Learning}} for {{Multiobjective Optimization}}.
\newblock {\em IEEE Transactions on Cybernetics}, 51(6):3103--3114, June 2021.

\bibitem{yaoEvolvingArtificialNeural1999}
Xin Yao.
\newblock Evolving artificial neural networks.
\newblock {\em Proceedings of the IEEE}, 87(9):1423--1447, September 1999.

\bibitem{kuremotoSearchHeuristicsOptimization2020}
Takashi Kuremoto, Takaomi Hirata, Masanao Obayashi, Kunikazu Kobayashi, and Shingo Mabu.
\newblock Search {{Heuristics}} for the {{Optimization}} of {{DBN}} for {{Time Series Forecasting}}.
\newblock In Hitoshi Iba and Nasimul Noman, editors, {\em Deep {{Neural Evolution}}: {{Deep Learning}} with {{Evolutionary Computation}}}, pages 131--152. {Springer Singapore}, {Singapore}, 2020.

\bibitem{rawalDiscoveringGatedRecurrent2020}
Aditya Rawal, Jason Liang, and Risto Miikkulainen.
\newblock Discovering {{Gated Recurrent Neural Network Architectures}}.
\newblock In Hitoshi Iba and Nasimul Noman, editors, {\em Deep {{Neural Evolution}}: {{Deep Learning}} with {{Evolutionary Computation}}}, pages 233--251. {Springer Singapore}, {Singapore}, 2020.

\bibitem{salimans2017evolution}
Tim Salimans, Jonathan Ho, Xi~Chen, Szymon Sidor, and Ilya Sutskever.
\newblock Evolution strategies as a scalable alternative to reinforcement learning.
\newblock {\em arXiv preprint arXiv:1703.03864}, 2017.

\bibitem{yangEvolutionaryReinforcementLearning2022}
Peng Yang, Hu~Zhang, Yanglong Yu, Mingjia Li, and Ke~Tang.
\newblock Evolutionary reinforcement learning via cooperative coevolutionary negatively correlated search.
\newblock {\em Swarm and Evolutionary Computation}, 68:100974, 2022.

\bibitem{zengEvolutionaryJobScheduling2023}
Detian Zeng, Jun Zhan, Wei Peng, and Zengri Zeng.
\newblock Evolutionary job scheduling with optimized population by deep reinforcement learning.
\newblock {\em Engineering Optimization}, 55(3):494--509, March 2023.

\bibitem{zhongAcceleratingGeneticAlgorithm2022}
Rui Zhong, Enzhi Zhang, and Masaharu Munetomo.
\newblock Accelerating the {{Genetic Algorithm}} for {{Large-scale Traveling Salesman Problems}} by {{Cooperative Coevolutionary Pointer Network}} with {{Reinforcement Learning}}.
\newblock {\em arXiv preprint arXiv:2209.13077}, 2022.

\bibitem{shaoMultiObjectiveNeuralEvolutionary2023}
Yinan Shao, Jerry Chun-Wei Lin, Gautam Srivastava, Dongdong Guo, Hongchun Zhang, Hu~Yi, and Alireza Jolfaei.
\newblock Multi-{{Objective Neural Evolutionary Algorithm}} for {{Combinatorial Optimization Problems}}.
\newblock {\em IEEE Transactions on Neural Networks and Learning Systems}, 34(4):2133--2143, April 2023.

\bibitem{yang2024reducing}
Peng Yang, Laoming Zhang, Haifeng Liu, and Guiying Li.
\newblock Reducing idleness in financial cloud services via multi-objective evolutionary reinforcement learning based load balancer.
\newblock {\em SCIENCE CHINA Information Sciences}, pages~--, 2023.

\bibitem{xinyaoEvolutionaryProgrammingMade1999}
Xin Yao, Yong Liu, and Guangming Lin.
\newblock Evolutionary programming made faster.
\newblock {\em IEEE Transactions on Evolutionary Computation}, 3(2):82--102, July 1999.

\bibitem{t.kailathDivergenceBhattacharyyaDistance1967}
{T. Kailath}.
\newblock The {{Divergence}} and {{Bhattacharyya Distance Measures}} in {{Signal Selection}}.
\newblock {\em IEEE Transactions on Communication Technology}, 15(1):52--60, February 1967.

\bibitem{beyerEvolutionStrategiesComprehensive2002b}
Hans-Georg Beyer and Hans-Paul Schwefel.
\newblock Evolution strategies \textendash{} {{A}} comprehensive introduction.
\newblock {\em Natural Computing}, 1(1):3--52, March 2002.

\bibitem{mazyavkinaReinforcementLearningCombinatorial2021}
Nina Mazyavkina, Sergey Sviridov, Sergei Ivanov, and Evgeny Burnaev.
\newblock Reinforcement learning for combinatorial optimization: {{A}} survey.
\newblock {\em Computers \& Operations Research}, 134:105400, 2021.

\bibitem{khalilLearningCombinatorialOptimization2017}
Elias Khalil, Hanjun Dai, Yuyu Zhang, Bistra Dilkina, and Le~Song.
\newblock Learning {{Combinatorial Optimization Algorithms}} over {{Graphs}}.
\newblock In {\em Advances in {{Neural Information Processing Systems}}}, volume~30. {Curran Associates, Inc.}, 2017.

\bibitem{zhengCombiningReinforcementLearning2021}
Jiongzhi Zheng, Kun He, Jianrong Zhou, Yan Jin, and Chu-Min Li.
\newblock Combining {{Reinforcement Learning}} with {{Lin-Kernighan-Helsgaun Algorithm}} for the {{Traveling Salesman Problem}}.
\newblock {\em Proceedings of the AAAI Conference on Artificial Intelligence}, 35(14):12445--12452, May 2021.

\bibitem{crama2005local}
Yves Crama, Antoon~WJ Kolen, and EJ~Pesch.
\newblock Local search in combinatorial optimization.
\newblock {\em Artificial Neural Networks: An Introduction to ANN Theory and Practice}, pages 157--174, 2005.

\bibitem{skinderowiczImprovingAntColony2022}
Rafa{\l} Skinderowicz.
\newblock Improving {{Ant Colony Optimization}} efficiency for solving large {{TSP}} instances.
\newblock {\em Applied Soft Computing}, 120:108653, May 2022.

\end{thebibliography}

\end{document}